\documentclass[letterpaper, 10 pt, conference]{ieeeconf}  % Comment this line out if you need a4paper

\IEEEoverridecommandlockouts                              % This command is only needed if 
\usepackage[pdftex]{graphicx}
\usepackage{multirow}
\usepackage{amsmath}
\usepackage{mathtools}
\usepackage{amssymb, bm}
\usepackage{amsfonts}
\usepackage[ruled,vlined]{algorithm2e}
\usepackage[dvipsnames]{xcolor}
\usepackage{url}

\usepackage{balance}

\graphicspath{{figures/}}

\title{\LARGE \bf
Distributed Mission Planning of Complex Tasks for Heterogeneous Multi-Robot Teams
}

\author{Barbara Arbanas Ferreira$^{1}$, Tamara Petrovi{\'{c}}$^{1}$ and Stjepan Bogdan$^{1}$% <-this % stops a space
     \thanks{$^{1}$Authors are with University of Zagreb, Faculty of Electrical Engineering and Computing, Laboratory for Robotics and Intelligent Control Systems (LARICS), Unska 3, Zagreb 10000, Croatia;
        {\tt\small barbara.arbanas@fer.hr}}%
     \thanks{This work has been supported by EU-H2020 CSA project AeRoTwin - Twinning coordination action for spreading excellence in Aerial Robotics, grant agreement No. 810321. The work of B. Arbanas Ferreira was supported in part by the “Young Researchers' Career Development Project–Training of Doctoral Students” of the Croatian Science Foundation funded by the European Union from the European Social Fund.}% <-this % stops a space
}

\begin{document}

\maketitle

%%%%%%%%%%%%%%%%%%%%%%%%%%%%%%%%%%%%%%%%%%%%%%%%%%%%%%%%%%%%%%%%%%%%%%%%%%%%%%%%
\begin{abstract}

In this paper, we propose a distributed multi-stage optimization method for planning complex missions for heterogeneous multi-robot teams. This class of problems involves tasks that can be executed in different ways and are associated with cross-schedule dependencies that constrain the schedules of the different robots in the system. The proposed approach involves a multi-objective heuristic search of the mission, represented as a hierarchical tree that defines the mission goal. This procedure outputs several favorable ways to fulfil the mission, which directly feed into the next stage of the method. We propose a distributed metaheuristic based on evolutionary computation to allocate tasks and generate schedules for the set of chosen decompositions. The method is evaluated in a simulation setup of an automated greenhouse use case, where we demonstrate the method's ability to adapt the planning strategy depending on the available robots and the given optimization criteria.

\end{abstract}


%%%%%%%%%%%%%%%%%%%%%%%%%%%%%%%%%%%%%%%%%%%%%%%%%%%%%%%%%%%%%%%%%%%%%%%%%%%%%%%%

\section{INTRODUCTION}
\label{sec:intro}

Cooperative multi-robot systems (MRS) have received much attention in recent decades \cite{survey2013,survey2019}. The great interest in these systems stems both from the considerable difficulty of establishing intelligent, coherent control of joint missions and from the many advantages that MRS bring. Compared to a single robot, MRS are able to leverage the strengths of the participating robots to establish a more robust system that is more resilient to various disturbances, such as robot or sensor failures. Furthermore, the introduction of heterogeneity, where each robot has different capabilities, leads to further interesting implications for the control system and allows for interesting collaborative behavior between robots \cite{survey-heterogeneous}.

Multi-robot systems have been studied from various aspects over the years. Our research focuses on the coordination and planning of cooperative missions for heterogeneous MRS. In this area, the development of a robust control architecture, communication and mission planning are the main problems discussed and solved in the literature \cite{survey2019}. In this paper, we focus on the problems of mission decomposition selection (the question of \emph{what do we do?}), task allocation (the question of \emph{who does what?}), and task scheduling (the question of \emph{how to arrange the tasks in time?}) of missions for MRS, which are often summarized under the common term \emph{mission (task) planning} \cite{Zlot2006}.

The missions we model in this paper fall into the class of problems CD[ST-MR-TA] defined in the taxonomy in \cite{korsahTaxonomy}. These missions involve tasks that may require execution by more than one robot (MR, multi-robot tasks), and robots may only perform one task at a time (ST, single-task robots). The task allocation and scheduling procedure considers both current and future assignments (TA, time-extended assignment). In terms of complexity, these tasks include complex task dependencies (CD), where each task can be achieved in multiple ways. The class CD also entails cross-schedule dependencies (XD), where various constraints relate tasks from plans of different robots.

To represent multi-robot missions, we use a hierarchical task model inspired by the language Task Analysis, Environment Modeling, and Simulation (T\AE MS) \cite{taems1999}. Its main premise is \emph{task decomposition}, where large and potentially complex tasks are incrementally decomposed into simpler ones, down to the level of actionable tasks (\emph{actions}). This tree-like hierarchical structure provides a good overview of the mission and the relationships between tasks, and greatly simplifies mission definition. Furthermore, the rich expressiveness of the mission formulation allows the definition of intricate task relations, and thus applicability in different domains.

In this paper, missions represented as large task hierarchies are subjected to a two-stage hierarchical optimization procedure. In the first step, we perform a fast and efficient heuristic search of the mission tree that finds several promising alternative ways to execute the mission (task decomposition selection procedure). Then, a task allocation and scheduling procedure \cite{ferreira2021distributed} is applied to several best-ranked alternatives to generate schedules for the given problem. Based on the given criteria, the best overall solution is output as the final schedule that best satisfies the mission objective.

The main contribution of this paper is the proposal of a fast and efficient distributed method for planning complex missions for heterogeneous MRS. The proposed multi-stage optimization approach provides a domain-independent solution to the given problem and can be readily applied to many areas of robotics research that involve cooperative robot teams. The method can adapt the planning strategy and select the appropriate tasks to execute, depending on the available robots and the given optimization criteria. In the current literature, there are not many approaches that attempt to generalize the planning procedure for generic tasks of class CD for heterogeneous multi-robot teams.

The above contributions are disseminated in the paper as follows. In the next section, we summarize current approaches to MRS coordination and mission planning and position our work in the state of the art. In Section \ref{sec:task_modeling}, we introduce the hierarchical task model used, which forms the basis for the proposed approach. Then, in Section \ref{sec:solution_approach}, we outline the multi-stage metaheuristic optimization approach used. In Section \ref{sec:results}, the performance of the proposed approach is analyzed using simulations of a use case of a robotized greenhouse. Finally, conclusions and plans for future work are presented in Section \ref{sec:conclusion}.

\section{RELATED WORK}
\label{sec:related_work}

Coordinating heterogeneous multi-robot teams requires precise high-level task planning and robust and efficient coordination mechanisms, and the literature offers many approaches to address this problem. From a control architecture perspective, these solutions can be divided into two groups -- centralized and distributed \cite{survey2019}. Although centralized architectures often produce optimal or near-optimal plans due to their global viewpoint \cite{Schillinger2018,Gombolay2018,Srinivasan2018}, distributed architectures typically exhibit better reliability, flexibility, adaptability, and robustness \cite{Floriano2019,Mitiche2019,Otte2017}, even if the solutions they provide are often suboptimal. % This makes them particularly suitable for field deployments in dynamic environments where a larger number of robots are used, even if the solutions they provide are often suboptimal.

Many of the current approaches rely on off-the-shelf automated reasoners based on, for example, Linear Temporal Logic (LTL) \cite{Schillinger2018,Srinivasan2018,DeCastro2018}. Although these solutions show significant contributions to the theoretical synthesis of correct-by-design controllers, they often suffer from intensive computational problems as well as the inability to quantify planner objectives and define complex task relationships.

Probabilistic multi-robot coordination approaches based on decentralized, partially observable Markov decision processes have also been studied \cite{Floriano2019,Morere2017,Pajarinen2017}. The advantage of this approach is its inherent suitability for uncertain environments; however, the scalability problem makes them unsuitable for real-world applications with multiple robots and complex tasks. Although some valuable online solutions have been reported \cite{Garg2019}, the problems that this approach can address are still relatively simple from a high-level mission planning perspective.

Some of the best known distributed solutions to the multi-robot mission planning problem are auction- and market-based approaches \cite{Otte2017,Nunes2017,Choi2009}. They usually solve the task allocation problem, where robots use bidding mechanisms for simple tasks that they assign to each other. However, the partial ordering between tasks and the tight coupling underlying our cooperative missions are often not considered. More recently, in \cite{Mitiche2019,Nunes2017}, the authors addressed the problem of precedence constraints in iterative auctions for the problem class XD.

On the other side of the spectrum, various optimization-based methods attempt to solve the task planning problem. They range from exact offline solutions \cite{Korsah2011} to heuristic approaches such as evolutionary computation and other AI optimization methods \cite{Mouradian2017,Jevtic2012,Luo2015}. In the former, the XD[ST-MR-TA] class problem is modeled as an instance of a Mixed-Integer Linear Programming (MILP) problem and solved using offline solvers or well-known optimization methods. Although optimal, the method is computationally expensive and lacks reactivity in dynamic environments.

% CD approaches
All of the above approaches have rarely attempted to solve the problem class CD that involves task allocation and scheduling and task decomposition selection. In our previous work \cite{Arbanas2018}, we attempted to solve the problem by adapting the Generalized Partial Global Planning (GPGP) framework \cite{Decker1995}, which specifies coordination mechanisms to enable cooperative behavior. Again, missions are represented in hierarchical tree structures, but a different mission planning strategy is used. In contrast to this approach, the GPGP-based solution solves allocation and scheduling problems separately. A greedy function assigns each task to the robot with the best score for that task, ignoring the impact on other task assignments. The scheduling procedure is performed locally for each robot using a genetic algorithm. The method was evaluated on a case of a symbiotic aerial-ground robotic team for autonomous parcel transport.

In \cite{Motes2020}, the authors present a multi-robot task and motion planner for multiple decomposable tasks with sequentially dependent subtasks. The planner is tailored to a case where tasks can be divided into subtasks that must be completed in a predefined order. The method is evaluated on a transportation task in a lake environment where the land and water robots can exchange a payload at dock locations. Each task in this example had three possible decompositions. Compared to our proposed method, this approach can only handle very simple and highly constrained tasks.

In \cite{Zlot2006}, the authors incorporate complex tasks into multi-robot task markets by including task tree auctions. Instead of trading contracts for simple tasks, task trees are offered in auctions. Complex tasks are specified as loosely coupled tasks connected by the logical operators \textit{AND} and \textit{OR}, similar to our mission specification. The method was tested with both centralized and decentralized setup on a reconnaissance mission that required coverage of multiple areas. The tree auction method outperformed all single-stage auction task allocation algorithms.

The advantage of our solution over the other state-of-the-art methods for CD in the literature lies in its generic approach, which ensures applicability to many different domains, without major intervention in the method itself. Moreover, we are able to specify complex optimization criteria for large mission structures with intricate task relationships. By combining a fast heuristic task decomposition selection method with a metaheuristic task assignment and scheduling optimization procedure, we can quickly search a large solution space and find efficient schedule configurations.
\section{HIERARCHICAL TASK REPRESENTATION}
\label{sec:task_modeling}

\begin{figure}[htb]
    \centering
    \includegraphics[width=.92\linewidth]{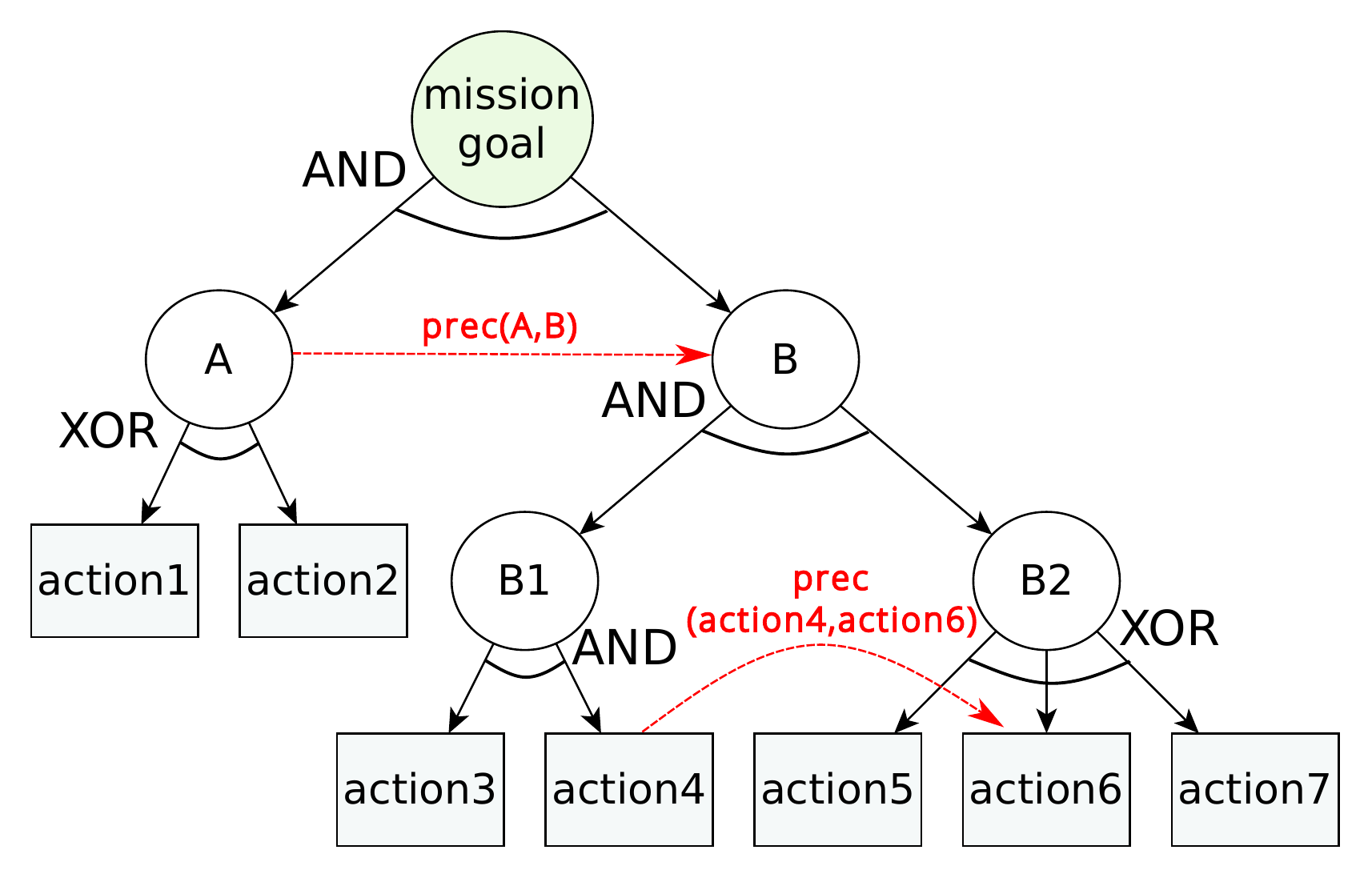}
    \caption{An example of a hierarchical task structure. Task nodes are represented by circles and action nodes by rectangles.}
    \label{fig:tree-example}
\end{figure}

Our task representation is based on T\AE MS \cite{taems1999}, a framework for representing large task hierarchies that allows the definition of simple and complex relations between tasks and temporal constraints on their execution. The mission tree contains action nodes that correspond to real, actionable robot behaviors, and task nodes that combine action and task nodes into a meaningful structure as defined by the mission objective. The tree-like structure enables the definition of multiple task levels, all contributing to the root-level task that represents the overall mission goal. An example of the hierarchical task structure is shown in Fig. \ref{fig:tree-example}.

Sets of actions and tasks are defined as $A$ and $T$, respectively. Each $a \in A$ can be performed by one or more robots. If we denote the set of robots as $R = \{1, \ldots, m\}$, we can specify the set of actions that robot $i$ can perform as $A_i$, and the set of tasks that robot $i$ can contribute to as $T_i$. Note that redundancy is possible, so in general $A_i \cap A_j$ and $T_i \cap T_j$ may not be empty sets, for $i \neq j, i, j \in R$.

Various task relations specify the effects of task execution on other tasks and resources. The most commonly used task relations that directly affect task order in the final schedule are \emph{precedence constraints}. They enforce the pairwise order of tasks or actions in the task structure. Formally, we define precedence constraints as $prec(a, b), a, b \in T \cup A$ for two tasks (or actions) $a$ and $b$. This relation defines that task $a$ must be executed before task $b$ is started. In the illustrative example in Fig. \ref{fig:tree-example}, the precedence constraints are shown as red dashed arrows.

Each task is quantitatively described in three dimensions: quality, duration, and cost. Quality is an abstract concept that depends on the problem domain and implies the contribution of a task to the achievement of the overall goal. Duration represents the time required to perform a particular task, and cost is the cost incurred to perform the task (which can be energy expenditure, financial cost, resources consumed, etc.).

To evaluate the tasks, each $a \in A$ is assigned a triple $(q_a(i), d_a(i), c_a(i))$, where $q_a(i)$ is the action quality, $d_a(i)$ is the duration, and $c_a(i)$ is the action cost when performed by robot $i \in R$. The action quality is determined a-priori by the system designer. Each robot estimates the duration and cost of a future action based on the current state of the system and their capabilities. The outcome of each task $t \in T$, $(q_t, d_t, c_t)$, is determined using the quality accumulation function $Q : T \rightarrow \mathbb{R}_0^3$, which describes how subtasks contribute to the quality of a higher-level task. In general, the function $Q$ can have any user-defined form.

In this paper, we use two functions corresponding to the logical operators \emph{\{AND, XOR\}}. The function \emph{AND} specifies the task decomposition of a task such that all subtasks of a task must be executed for it to acquire quality. The quality of the parent task is calculated as the \emph{sum} of all subtask qualities. The \emph{XOR} function of a composite task requires the execution of \emph{exactly one} subtask. The quality of the parent task is equal to the quality of the chosen subtask. For a given solution, only tasks that achieve a quality greater than $0$ are considered accomplished.

\section{MULTI-STAGE METAHEURISTIC OPTIMIZATION FOR COMPLEX MISSIONS}
\label{sec:solution_approach}

The goal of the planning procedure is to create schedules for all robots in the system based on the mission specified in terms of a previously established hierarchical task model. The planning process is in search of a correct solution that most satisfies the overall objective function. Since the mission planning problem is of class CD, our approach involves a multi-stage optimization procedure. First, a heuristic task decomposition selection generates possible task decompositions, and the several best ones are subjected to a metaheuristic allocation and scheduling procedure that outputs the final schedule.

\subsection{Heuristic task decomposition selection procedure}

The problem of task decomposition selection involves finding a subset of actions and tasks to be performed that are most promising to provide near-optimal schedules. In this step of the procedure, a heuristic tree search algorithm is used to quickly generate alternative subsets of tasks that satisfy the mission objective.

We define a \textit{task alternative} $alt(t)$, $alt(t) \subseteq A$, $t\in T$ as an unordered set of all actions whose execution leads to the completion of the task $t$. The sizes (cardinal numbers) of the task alternative sets for different tasks in the mission plan depend on the structure of the mission tree and the relationships between nodes. For highly constrained missions, the cardinal numbers are generally small (i.e., $O(1)$). On the other hand, for missions without any node interrelations, the combinatorial explosion can lead to a factorial size complexity of the task alternative generation procedure.

\begin{algorithm}[h!]
  \SetAlgoLined
  \SetKwInOut{Input}{input}
  \SetKwInOut{Output}{output}
  \SetKwInOut{Param}{parameter}
  \SetKwFunction{FAlt}{generate\_alternatives}
  \SetKwProg{Fn}{Function}{:}{}
  \Param{$\mu$ -- max alternative number for a single task}
  \Input{$tree$ -- hierarchical task tree specifying the mission}
  \Input{$criteria$ -- evaluation criteria}
  \Output{$task\_alt$ -- alternative decompositions of task $task$}
  \Fn{\FAlt{$task$}}{
        \tcc{recursion stopping criteria}
        \If{$task \in A$}{
            \Return [$task$]\;
        }
        \tcc{recursively generate task alternatives}
        $alt \gets$ []\;
        \For{$subtask \in tree.subtasks(task)$}{
            $alt \gets alt\ \cup $ generate\_alternatives($subtask$)\;
        }
        \tcc{combine subtask alternatives based on function $Q$}
        \Switch{$tree.Q(task)$}{
            \Case{$AND$}{
                $task\_alt \gets $ cartesian\_product($alt$)]\;
            }
            \Case{$XOR$}{
                $task\_alt \gets $ [[$task \cup alt^*$] for $alt^* \in alt$]\;  
            }
        }
        \tcc{pruning procedure}
        \If{$|task\_alt| > \mu$}{
            $sc \gets $ evaluate\_alternative($task\_alt$, $criteria$)\;
            $top \gets $ index\_of\_max\_n\_elements($sc, \mu$)\;
            $task\_alt \gets [task\_alt(i), i \in top$]\;
        }
        \Return[$task\_alt$]
  }
 \caption{Heuristic alternative generation procedure.}
 \label{alg:alt-generation}
\end{algorithm}

The process of generating task alternatives starts at the action nodes of the mission tree and builds up recursively, eventually ending at the root of the tree, as outlined in Algorithm \ref{alg:alt-generation}. To tame a potential combinatorial explosion, the procedure uses a method of focusing the solution search by pruning the worst partial results at each step of the process to make the problem tractable. During this procedure, the robots use estimated values for quality, duration, and cost of actions, which are determined as the average of these values for all robots that can perform the action, as follows:
\begin{equation}
    (\overline{q}_a, \overline{d}_a, \overline{c}_a)= \sum_{i \in \rho_a}{\frac{(q_a(i), d_a(i), c_a(i))}{|\rho_a|}}, a \in A.
\end{equation}
$\rho_a$ defines the set of robots that can perform action $a$, $\rho_a = \{i, i \in R, a \in A_i\},\ \forall a \in A$. $|\rho_a|$ stands for the cardinality of the set $\rho_a$.

Finally, the score for each task alternative $alt(t)$ is computed based on the expected values of actions $(\overline{q}_a, \overline{d}_a, \overline{c}_a)$, and given the quality accumulation function $Q$ and the defined tree structure. If we specify the alternative outcome values as $(q_{alt}, d_{alt}, c_{alt})$, our simplified objective function (score) for an alternative is defined as 
\begin{equation}
	\label{eq:alter_score}
	sc(q_{alt}, d_{alt}, c_{alt}) = \alpha q_{alt} - \beta d_{alt} - \gamma c_{alt},\ \alpha, \beta, \gamma \in \mathbb{R},
\end{equation}
where $\alpha + \beta + \gamma = 1$, and they represent the importance weighting of each specific criterion. Based on these factors, the planning strategy adapts and selects the appropriate tasks to execute.

Based on this score function, we are able to define the importance of each problem parameter in the task decomposition selection. As a result of this process, the robots are given a set of alternative ways to achieve the mission goal (root task). %These alternatives are a direct input for the next phase of the optimization, allocation and scheduling procedure.

\subsection{Distributed task allocation and scheduling procedure}

The task allocation and scheduling procedure considers a problem where a team of heterogeneous robots $R = \{1,\ldots,m\}$ is available to perform a collection of simple single-agent tasks (\emph{actions}) $A = \{1,\ldots,n\}$. In our multi-stage optimization scheme, the set of actions to be scheduled in this phase is provided as a selected set of alternatives for the root task, $alt(root)$. A solution to the described problem is a set of time-related actions (\emph{schedule}) for all robots. Formally, the schedule $s_i$ for each robot $i \in R$ is defined as $s_i = \{(a, a^s, a^f)\ \forall a \in S_i\}$, where $S_i$ is the set of actions assigned to robot $i$, and $a^s (a^f)$ are the start (finish) times of action $a$.

All solutions found must adhere to the \emph{precedence constraints} specified by the mission structure, which are defined as follows. If action $a \in A$ must complete before action $b \in A$ begins, a constraint is generated as $prec(a, b)$. This constraint enforces $a^f < b^s$, where $a^f$ and $b^s$ specify the times at which action $a$ completes and $b$ begins.

\subsubsection{Problem modeling}

In solving this problem, we use modeling of the defined problem as a form of Multi-Depot Vehicle Routing Problem (MDVRP) with precedence constraints. In essence, MDVRP is a problem in which vehicles with limited payloads must pick up or deliver items at different locations. The items have a quantity, such as weight or volume, and the vehicles have a maximum capacity that they can carry. The problem is to pick up or deliver the items at the lowest cost without exceeding the vehicle capacity. Given the many similarities between task planning and MDVRP, by modeling the task planning problem as a variant of the Vehicle Routing Problem (VRP), we can apply many optimization techniques already available for VRP problems to our problem. Moreover, the MDVRP representation generalizes the problem so that the solution can be readily applied to many different domains. A similar modeling was proposed in \cite{Korsah2011}, where the task planning problem refers to the Dial-a-Ride Problem (DARP), a variant of VRP with pickup and delivery.

\begin{figure}[htb]
    \centering
    \includegraphics[width=\linewidth]{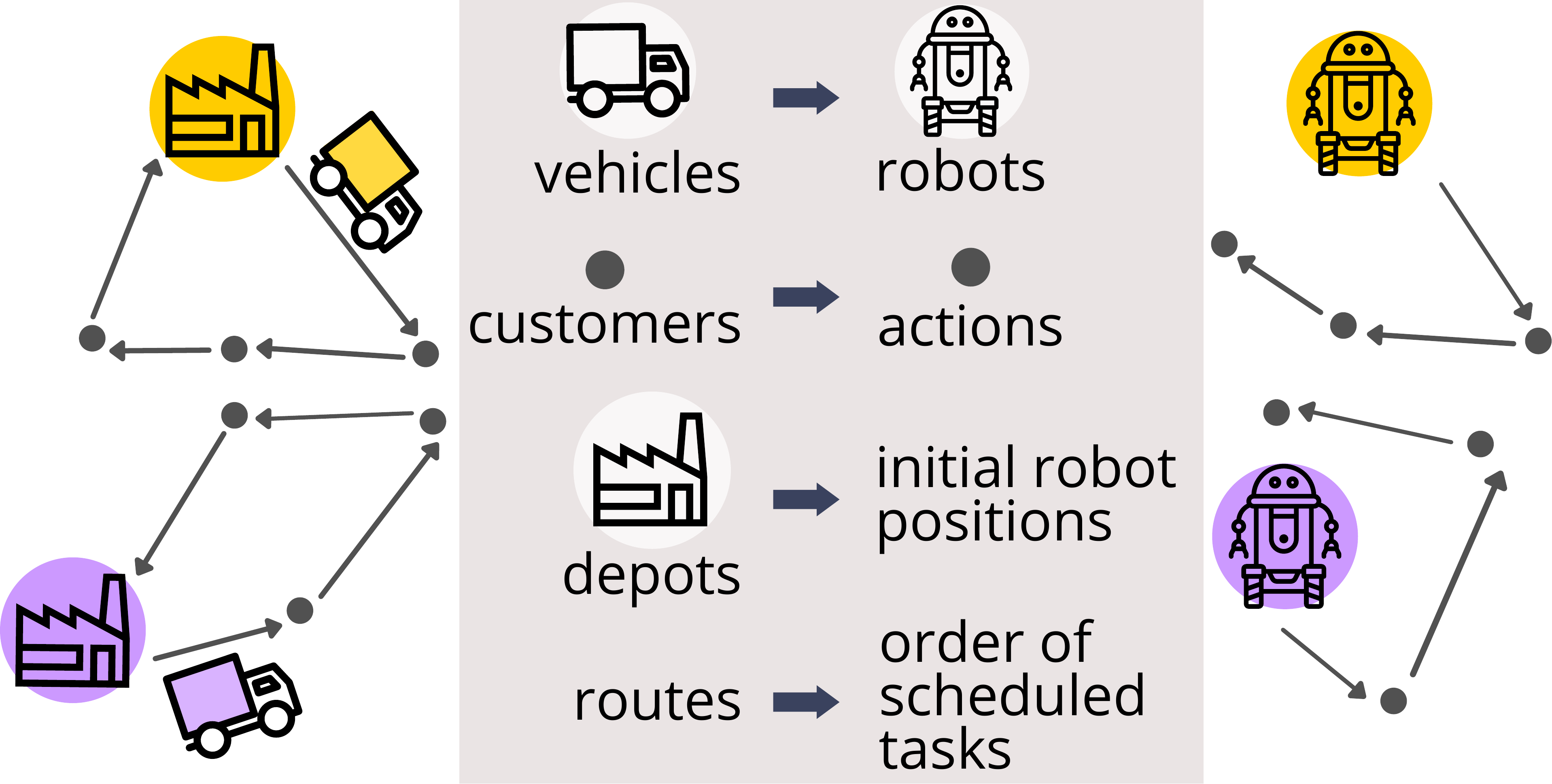}
    \caption{An illustration of the relationship of the MDVRP model to the task planning paradigm. The diagram shows the direct relation between concepts in VRP and task planning problems.}
    \label{fig:vrp-planning}
\end{figure}

In relating the task planning problem to the MDVRP model, we associate the basic VRP concepts directly with the task planning paradigm. As shown in Figure \ref{fig:vrp-planning}, the idea of a depot in VRP problems is directly associated with the initial position of the robot, and the vehicle in VRP represents a robot itself. The concept of customer and customer demand is applied to actions in task planning and the cost of each action, respectively. Consequently, routes as solutions to VRP problems represent the order of actions in the final robot schedules in the task planning model. More details on the mission modeling are provided in our previous work \cite{ferreira2021distributed}.

\subsubsection{Distributed metaheuristic algorithm for XD task planning problems}

As a solution to the stated problem, we employ multi-objective optimization with a form of distributed genetic algorithm with mimetism and knowledge sharing. This approach, which uses methods of distributed evolutionary computation, can quickly generate near-optimal solutions and thus work online while achieving good scalability properties. The method we use is inspired by the Coalition-Based Metaheuristic (CBM) algorithm \cite{Meignan2009}, with specific implementation to meet our problem requirements. In this section, we briefly outline the algorithm, and more details are available in our paper \cite{ferreira2021distributed}.

The algorithm behaves very similarly to Genetic Algorithm (GA) in that it uses the same solution representation in terms of chromosomes to which a set of genetic operators is applied. An important difference is that in CBM, the selection of the operator to apply is not completely stochastic, since this algorithm stores knowledge about its past actions and their effects in order to identify those that are more likely to lead to better solutions. Moreover, CBM is a distributed algorithm that runs on multiple agent nodes (in a robotic system, agents are robots). The robots can share their accumulated experience as well as the best solutions they have found. Therefore, each robot not only learns from its own experience, but can also exhibit mimetic behavior.

Inspired by an evolutionary process, solutions represented as chromosome contain genetic material (genotype) that defines a solution. For our problem, this refers to the assignment of actions to different robots and their order within the schedule. Each chromosome is associated with a phenotype that evaluates the genetic material and, in our case, generates schedules for task sequences based on the temporal properties of the tasks (task duration, time of transitioning between tasks).

\begin{figure}[htb]
    \centering
    \includegraphics[width=.9\linewidth]{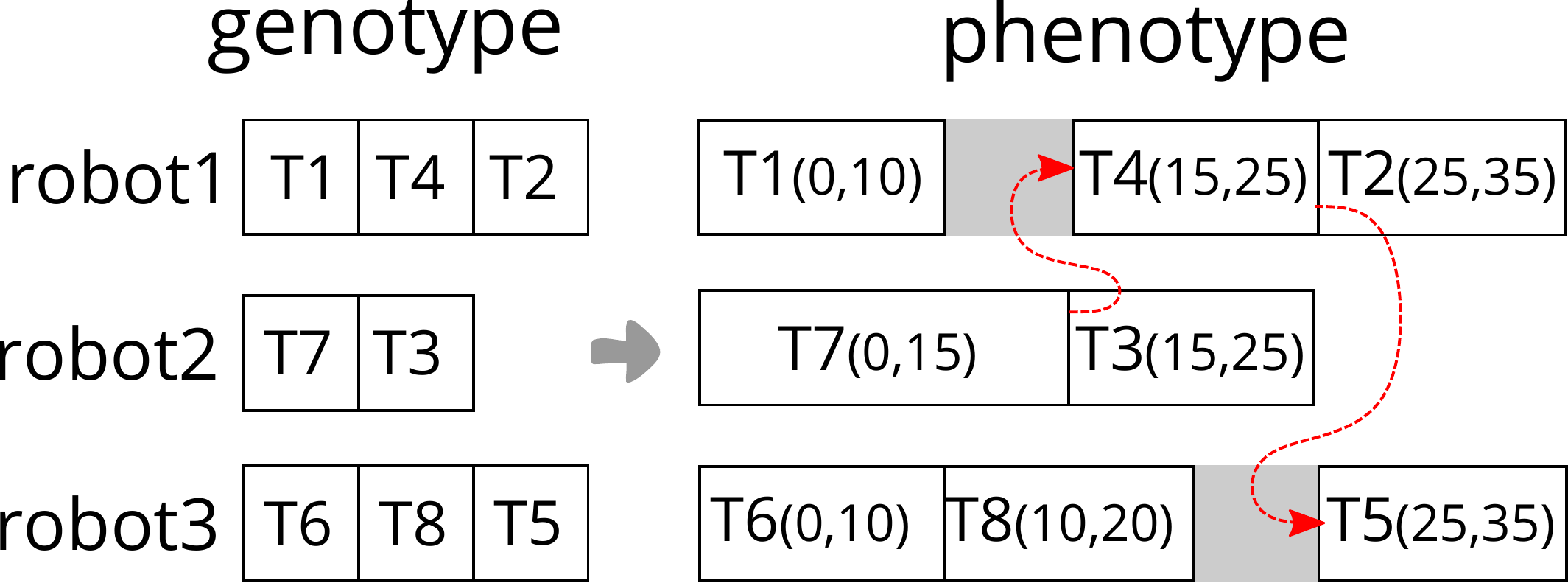}
    \caption{Solution representation -- chromosome genotype and phenotype. The tasks presented here are arbitrarily named generic tasks ($\{T1,\ldots,T8\}$). The idle times introduced in the schedules are a consequence of precedence constraints $prec(T7,T4)$ and $prec(T4,T5)$, since task $T4$ cannot start before task $T7$ finishes, and task $T5$ cannot start before the end of $T4$.}
    \label{fig:chromosome}
\end{figure}

An example of a chromosome and its genotype and phenotype is shown in Figure \ref{fig:chromosome}. The left side represents the genetic material of the given example solution, which contains the task assignments for the robots $\{robot1, robot2, robot3\}$ and their order. The genotype representation is maintained preserving the intra-schedule precedence constraints. The right figure represents the phenotype of the specified genotype. During phenotype generation, minimal idle times are inserted as needed to ensure consistency with the defined inter-schedule constraints. The phenotype represents the so-called semi-active schedule, where no left shift is possible in the Gantt graph. For any given sequence of robot operations, there is only one semi-active schedule \cite{Sprecher1995}. A major advantage of this type of solution design is faster exploration of the solution space, since all operators perform on a simpler genotype representation of the solution. The evaluation procedure renders the phenotype and evaluates the solutions found.

During the optimization procedure, we keep a population of solutions on which different genetic operators are performed. To evaluate solutions in a population, we apply a double-rank strategy which scores the solutions based on several criteria. Specifically, we use two criteria, makespan of the schedule and the total cost of executing the schedule. In the first part of the evaluation, we use a Pareto ranking procedure \cite{Kalyanmoy2001} that assigns ranks to all solutions based on the non-dominance property (i.e., a solution with a lower rank is clearly superior to solutions with a higher rank concerning all objectives). Therefore, the solutions are stratified into multiple ranks based on their ability to meet the optimization objectives. The second part of the evaluation function is the density function, which determines how similar the solution is to other individuals in the population. Finally, the rank and density scores are combined in the fitness of each solution $fitness \in \mathbb{R}_0$.

Several genetic operators guide the exploration of the solution space. We applied genetic crossover and mutation operators from the literature \cite{Pereira2009}, adapted them to the specifics of our problem. The implemented crossover operator is a version of Best-Cost Route Crossover (BCRC). Here, one route to be removed is selected for each of two parent chromosomes. The removed nodes are inserted into the other parent with the best insertion cost. As mutation operators, we use intra-depot and inter-depot swapping procedures, which select two random routes from the same (for intra-depot) or different (for inter-depot) starting position and swap a randomly selected action from one route to the other. Another mutation method is single action rerouting, where an action is randomly selected and removed from the existing route. The action is then inserted at the best feasible insertion point within the entire chromosome. The details of each operator are not included in this paper due to space limitations.

\section{SIMULATION RESULTS}
\label{sec:results}

\begin{figure}[htb]
    \centering
    \includegraphics[width=\linewidth]{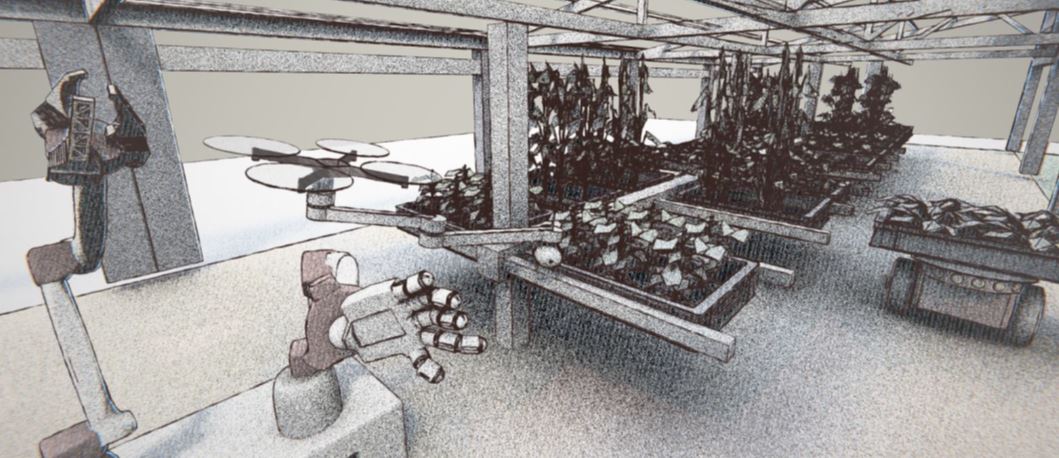}
    \caption[]{An illustration of a robotized greenhouse maintained by a heterogeneous robotic team\footnotemark.}
    \label{fig:specularia}
\end{figure}

To evaluate the performance of the proposed method, we developed a practical use case with problems of class CD. The application is based on an automated greenhouse and the scheduling of its daily maintenance tasks. The greenhouse structure is organized as a set of tables, comprising several plant containers representing growth units. In this example, each container holds a single plant that can be conveyed through the greenhouse using a UGV with a special mechanism for transporting plants. An illustration of such a system is given in Fig. \ref{fig:specularia}.

The UGV works in symbiosis with a stationary manipulator located at a workspace with four empty container holders. The manipulator can perform various operations on the plants once they are brought into the workspace. The design of a workspace with multiple slots allows batch operations on plants, speeding up some of the procedures. Naturally, not all tasks support batch processing to the same degree.

In addition to the stationary manipulator, we also envisioned a mobile manipulator consisting of a larger UGV with a robotic arm on board. This robot is capable of driving around the greenhouse and tending the plants directly. Note that each of the operations in this case must be performed from both sides of the table to take into account the entire plant.

\begin{figure}[htb]
    \centering
    \includegraphics[width=\linewidth, clip, trim=4.5cm 2cm 4cm 3.2cm]{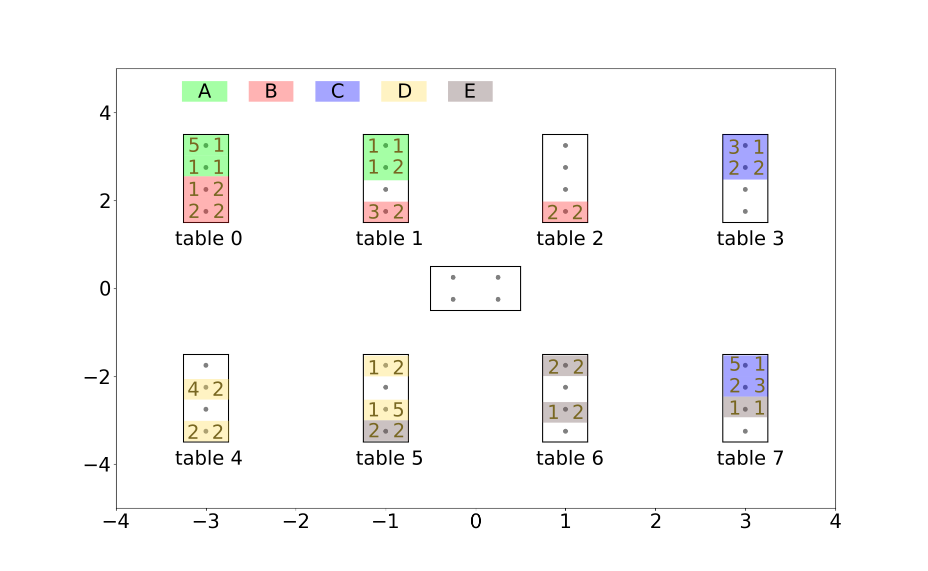}
    \caption{Greenhouse use case plant setup. In the illustration, the different batches of plants are marked in different colors, as indicated in the legend. The dimensions of the greenhouse (in meters) are shown on the x and y axes.}
    \label{fig:greenhouse_setup}
\end{figure}

\footnotetext{\url{http://specularia.fer.hr}}

The structure of the greenhouse we used for the simulations is shown in Fig. \ref{fig:greenhouse_setup}. The structure consists of eight tables, each with 4 plants. For the purpose of batch processing, the plants were clustered a-priori into five groups $\{A,B,C,D,E\}$ as shown in the figure. In this arrangement, there are a total of $20$ plants that need to be tended to. The numbers given in the table structures for each plant denote the number of unit operations that need to be performed on each side of the plant. They are used to estimate the total procedure duration. In our example, we specify the unit operation duration as $10s$. Therefore, using the example of the top plant in $table 0$, the total processing time on the left side is $50s$ and on the right side is $10s$.

\begin{figure}[htb]
    \centering
    \includegraphics[width=.95\linewidth]{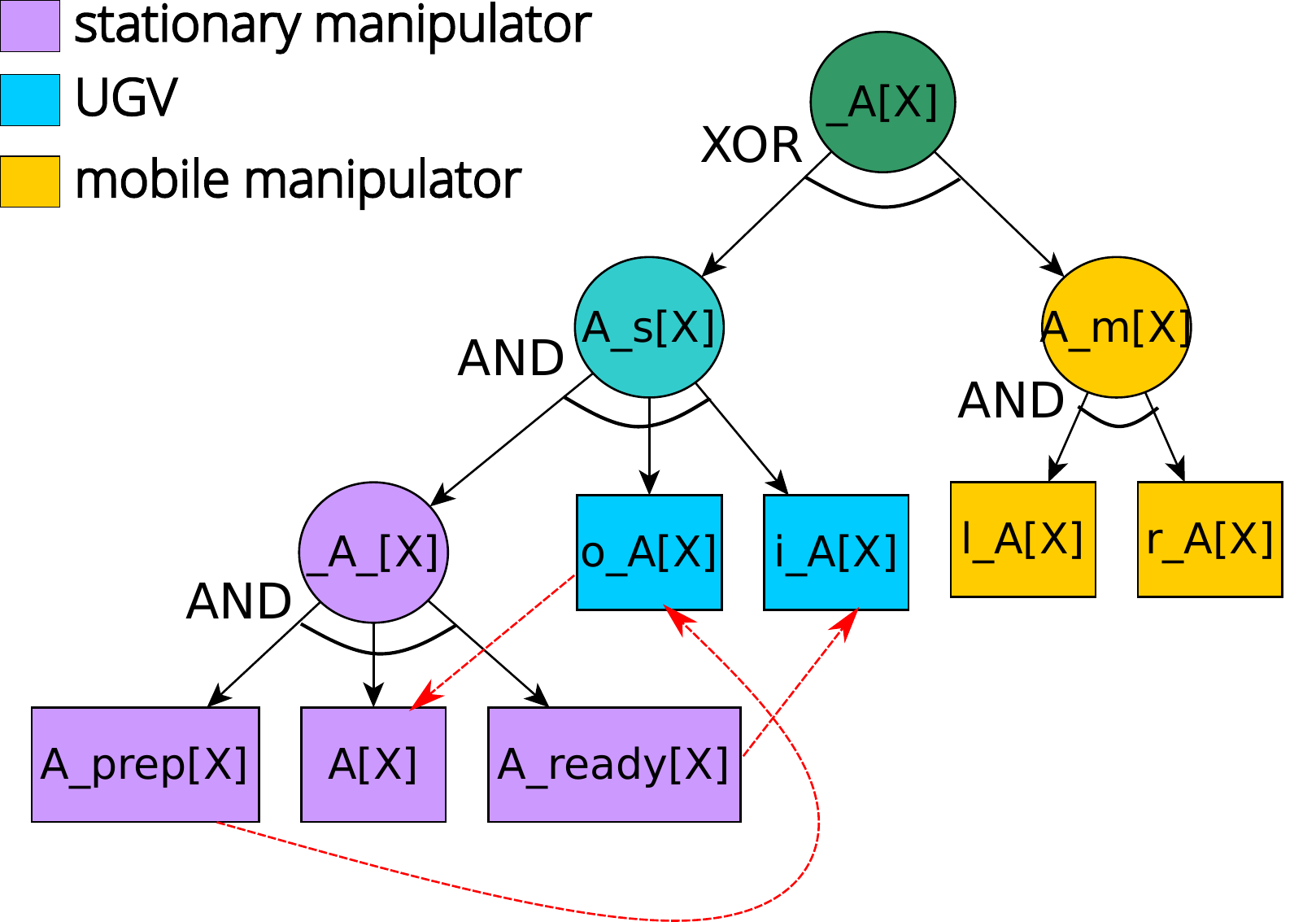}
    \caption{Hierarchical task structure for the task of processing a single plant in the system. $A$ is the plant type symbol and $X$ stands for a unique plant identifier, which is composed as \textit{"table.row.column"} of the plant in the greenhouse structure.}
    \label{fig:taems_template}
\end{figure}

The described mission is modeled as a hierarchy of tasks, as described earlier. A task tree for a single plant operation is shown in Fig. \ref{fig:taems_template}. At the root of the structure is a task representing the desired operation, and it is divided into two subtasks denoting two alternative ways of processing the plant. The left variant defines a stationary case where a UGV and a static manipulator perform the task together. The UGV has to deliver the plant to the workspace ($o\_A$) and return it when the task is finished ($i\_A$). For the stationary manipulator, we defined three tasks, $A\_prep$, $A$, and $A\_ready$. These tasks are executed sequentially, and the $prep$ and $ready$ tasks are used to synchronize the operations of the batch procedure with other plants of the group. On the other hand, there is an option where a mobile manipulator tends to the plant, and it includes tasks of the left ($l\_A$) and right ($r\_A$) processing. The full mission structure includes $20$ models of this structure, one for each plant, and they are associated with operator \textit{AND}.

\begin{table}[htb]
    \centering
    \caption{Simulation setups for the use case.}
    \label{tab:results}
    \scalebox{0.95}{
    \begin{tabular}{cccccc}
         \multirow{2}{*}{setup} & \multirow{2}{*}{\shortstack{problem\\class}} & \multirow{2}{*}{\shortstack{mobile\\manipulator}} & \multirow{2}{*}{\shortstack{stationary\\manipulator}} & \multirow{2}{*}{UGV} &  \multirow{2}{*}{\shortstack{makespan-cost\\importance}} \\
         & & & & & \\
         \hline
         1 & XD & - & 1 & 2 & - \\
         2 & XD & 2 & - & - & - \\
         3 & CD & 1 & 1 & 1 & 0-100 \\
         4 & CD & 1 & 1 & 1 & 50-50 \\
         5 & CD & 1 & 1 & 1 & 100-0 \\
    \end{tabular}
    }
\end{table}

\begin{figure}[htb]
    \centering
    \includegraphics[width=\linewidth]{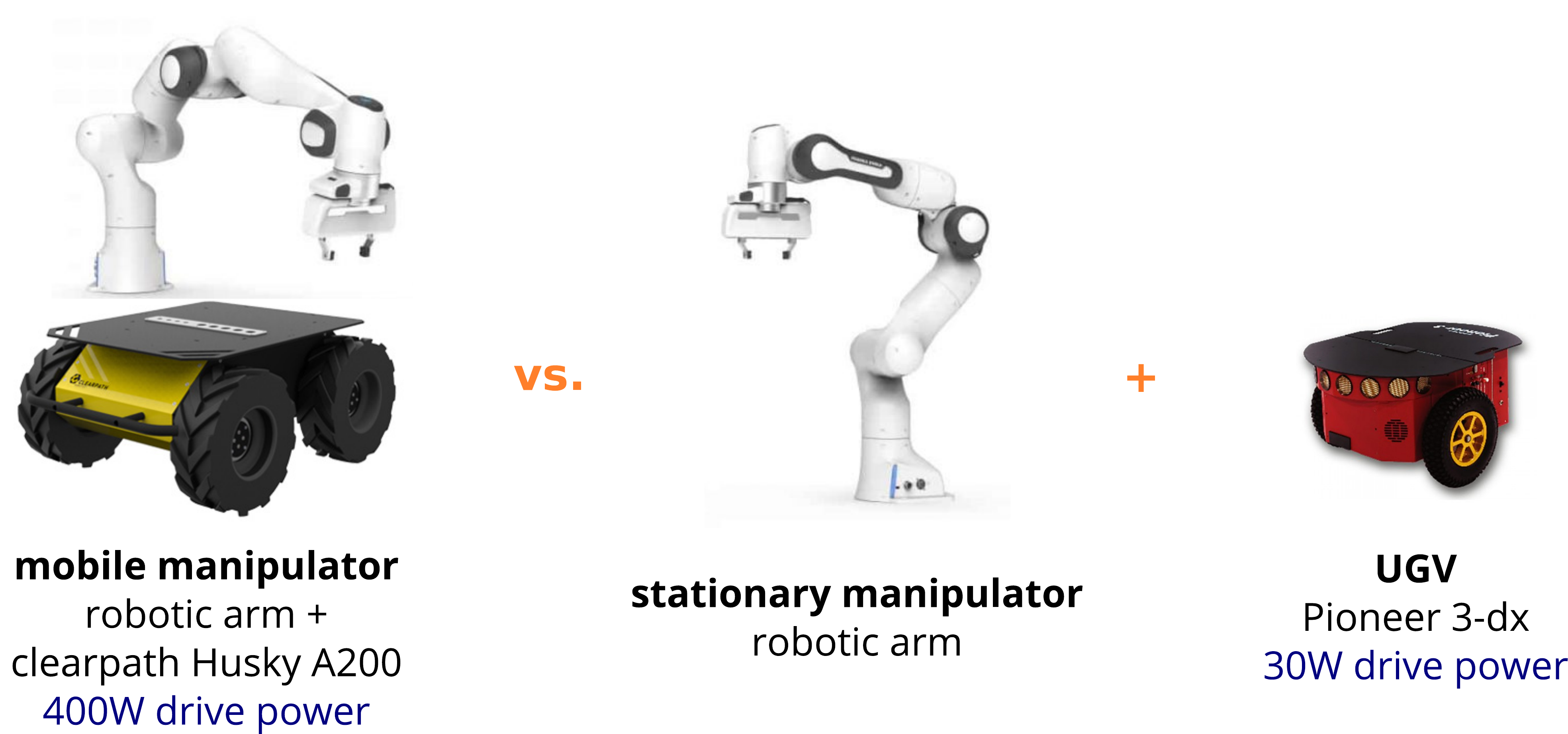}
    \caption{Robot team setups in the simulation scenarios. We compare the performance of a mobile manipulator with that of a stationary manipulator supported by UGVs. For complex missions, we allow the use of combinations of all three robots.}
    \label{fig:robots_usecase}
\end{figure}

We ran several simulations for the setups defined in Table \ref{tab:results}. The first two setups compare using a single stationary manipulator and two UGVs, versus using two mobile manipulators. For the next three setups, we used one of each of the three robot classes but varied the importance of mission makespan and cost in the evaluation procedure. To estimate the cost of each task, we assume a realistic set of robots, where for a UGV, we suggest a Pioneer 3-DX robot \cite{pioneer} with an estimated drive power of $30W$, while in the case of a UGV carrying a manipulator, we propose a more robust solution of a Clearpath Robotics Husky A200 \cite{husky} with a drive power of $400W$. An illustration of the robotic system configurations is given in Fig. \ref{fig:robots_usecase}. The maximum speeds for both robots are set to $0.5m/s$, taking into account the sensitive payload they have on board. Based on the duration of each task, we calculate the energy consumed in $kJ$ by multiplying the duration and the power demand. This is a simplified form of a cost function and serves the purpose of testing the planning system. For a more accurate model, the different power requirements of the robots depending on the actual battery charge should be considered.

\begin{figure}[htb]
    \centering
    \includegraphics[width=\linewidth]{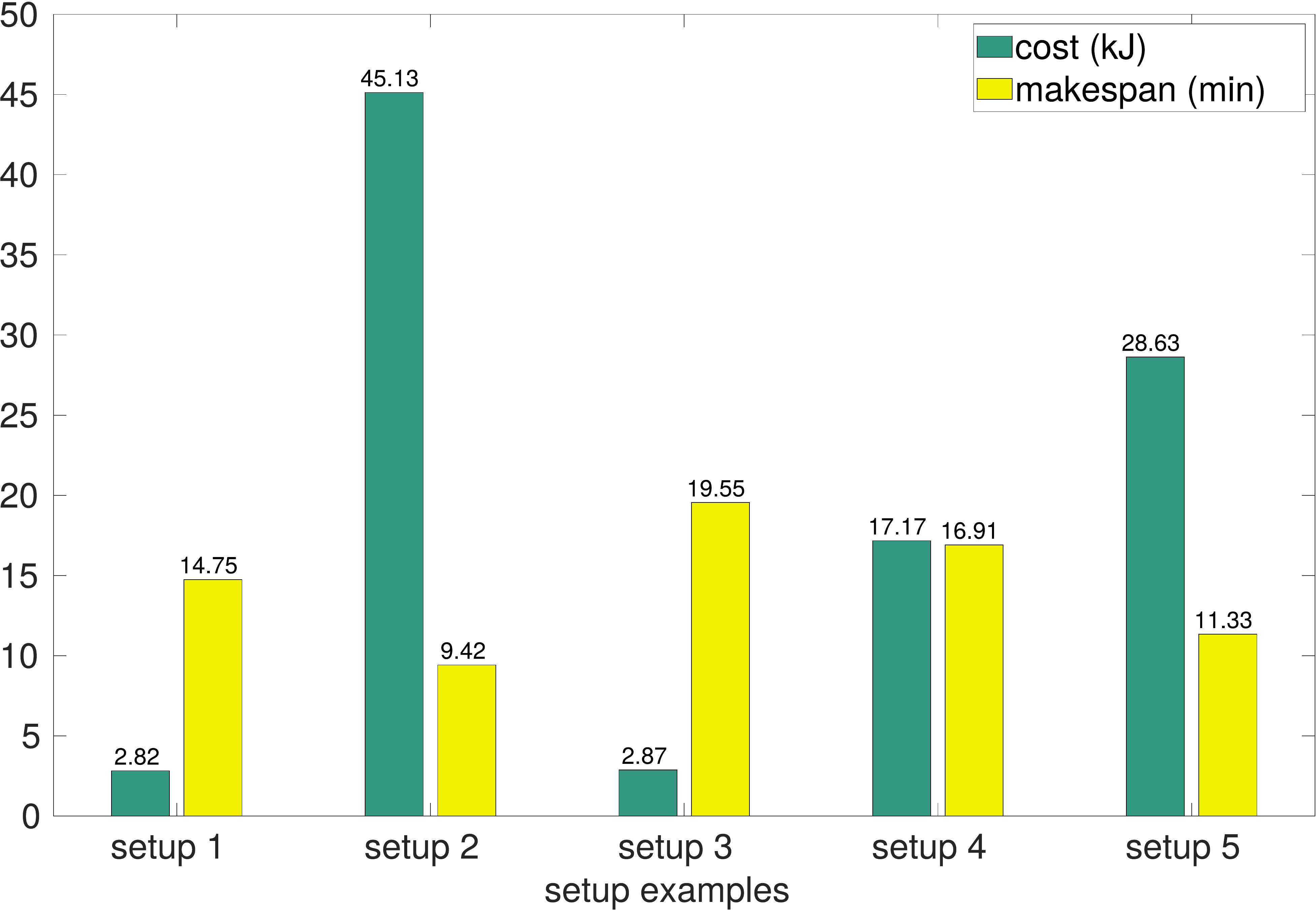}
    \caption{Makespan and cost of best found solutions for each setup.}
    \label{fig:results-plot}
\end{figure}

The results of the simulation runs for each setup are shown in Figure \ref{fig:results-plot}. There is a clear discrepancy in the ratio of makespan to cost for the first two configurations, where the setup with one stationary arm and two light UGVs consumes much less energy, with the mission time span increasing by $57\%$. On the other hand, it can be seen that the second configuration consumes $16$ times more energy for execution, although it is faster. To exploit the strengths of both robots, we consider the case of combining these two approaches in the following setups. The proposed mission decomposition selection procedure allows to choose the best way of mission execution considering a certain criterion. Here we analyze three extreme cases: the preference of cost savings in \textit{setup 3}, the equal importance of cost and makespan in \textit{setup 4}, and finally the preference of mission speed without considering cost in \textit{setup 5}.

The results for these setups demonstrate the ability of our proposed complex mission planning system to select appropriate mission decompositions given the robots available in the system and the specified criteria. In a case where the cost is evaluated higher, all tasks are selected to be executed in an energy-efficient manner, which corresponds to the left branch of the mission tree defined in \ref{fig:taems_template}. For the case where fast execution is required, right branches of the mission tree are selected for all available plants, which are handled by a mobile manipulator. For the middle case, the procedure outputs a solution that balances the tasks between these two options.

\section{CONCLUSION AND FUTURE WORK}
\label{sec:conclusion}

In this paper, we developed a distributed multi-stage optimization method for planning complex missions for heterogeneous multi-robot teams. The missions we model in this work fall into the class of complex dependency problems (CD), where each task has multiple ways to complete. %The class CD also includes cross-schedule dependencies (XD), where various constraints relate tasks from plans of different robots. 
Our solution in the first stage of optimization focuses on the task decomposition problem (determining a set of actions to perform). A fast heuristic tree search algorithm generates alternative subsets of tasks that satisfy the mission objective. In the next step, a multi-objective optimization using a distributed genetic algorithm with mimetism and knowledge sharing addresses the task allocation and scheduling problem. Using a simulated application example of a robotized greenhouse with a heterogeneous multi-robot team, we demonstrate the ability of our proposed complex mission planning system to select appropriate mission decompositions given the robots available in the system and the given criteria.

As future work, we are interested in extending the model to allow for more complex mission specifications in terms of task decomposition. We are also interested in analyzing the impact of agent or communication failures and finding ways to make the method more robust to real application problems. We also plan to test and evaluate the proposed method in an experiment in robotic greenhouse environment.




%%%%%%%%%%%%%%%%%%%%%%%%%%%%%%%%%%%%%%%%%%%%%%%%%%%%%%%%%%%%%%%%%%%%%%%%%%%%%%%%

\bibliographystyle{bibliography/IEEEtran}
\balance
\bibliography{bibliography/root.bib}

\end{document}